\documentclass[10pt,twocolumn,letterpaper]{article}

\usepackage{cvpr}              

\usepackage{graphicx}
\graphicspath{{figures/}}
\usepackage{amsmath}
\usepackage{amssymb}
\usepackage{booktabs}
\usepackage{multirow}
\usepackage[pagebackref,breaklinks,colorlinks]{hyperref}

\usepackage[capitalize]{cleveref}
\crefname{section}{Sec.}{Secs.}
\Crefname{section}{Section}{Sections}
\Crefname{table}{Table}{Tables}
\crefname{table}{Tab.}{Tabs.}

\def\cvprPaperID{12154} 
\def\confName{CVPR}
\def\confYear{2023}

\begin{document}

\title{That’s BAD: Blind Anomaly Detection by Implicit Local Feature Clustering}

\author{Jie Zhang$^1$
~~~~Masanori Suganuma$^1$
~~~~Takayuki Okatani$^{1,2}$\\
$^1$Graduate School of Information Sciences, Tohoku University ~~~~ $^2$RIKEN Center for AIP \\
{\tt\small \{jzhang,suganuma,okatani\}@vision.is.tohoku.ac.jp}
}
\maketitle

\begin{abstract}
Recent studies on visual anomaly detection (AD) of industrial objects/textures have achieved quite good performance. 
They consider an unsupervised setting, specifically the one-class setting, in which we assume the availability of a set of normal (\textit{i.e.}, anomaly-free) images for training.
In this paper, we consider a more challenging scenario of unsupervised AD, in which we detect anomalies in a given set of images that might contain both normal and anomalous samples. The setting does not assume the availability of known normal data and thus is completely free from human annotation, which differs from the standard AD considered in recent studies. For clarity, we call the setting blind anomaly detection (BAD). We show that BAD can be converted into a local outlier detection problem and propose a novel method named PatchCluster that can accurately detect image- and pixel-level anomalies. Experimental results show that PatchCluster shows a promising performance without the knowledge of normal data, even comparable to the SOTA methods applied in the one-class setting needing it. 
\end{abstract}

\section{Introduction}
\begin{figure}
  \centering
    \includegraphics[width=1\linewidth]{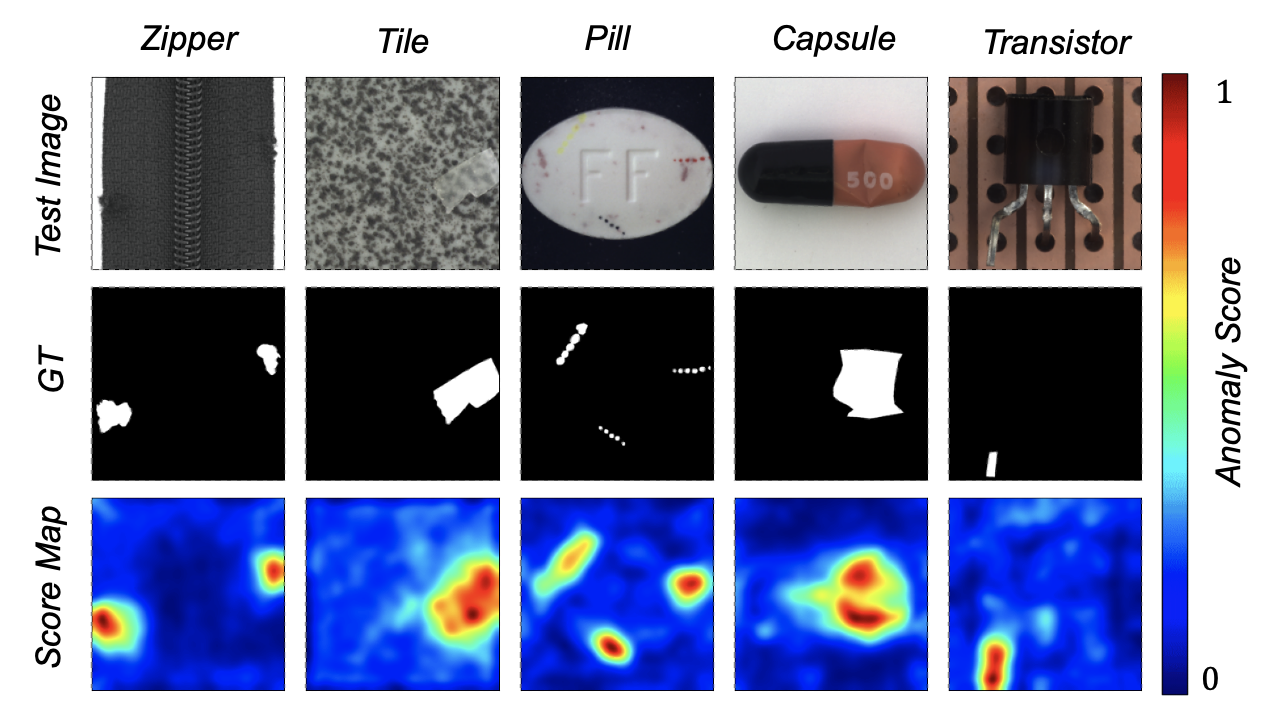}

\caption{Examples of industrial anomalies from the MVTec AD and the detection results of PatchCluster-25\% without using the training data.}
\label{bad_examples}
\end{figure}

In this paper, 
we consider visual anomaly detection and localization of industrial objects and textures.
Anomaly detection (AD) \cite{chandola2009anomaly,chalapathy2019deep,xia2015learning} is the task of detecting anomalous images or patterns that are out of the distribution of normal images or patterns. 
AD for industrial applications often requires distinguishing small differences between normal and anomalies \cite{bergmann2018improving,rudolph2021same,bergmann2022beyond}; see examples from a standard benchmark dataset, MVTec AD, in Fig.\ref{bad_examples}. 

As anomalies can appear with countless types, and the majority are usually the normal samples in a manufacturing line, the one-class unsupervised setting has drawn the most attention from the research community. This setting assumes a set of anomaly-free images for training, which are selected by human experts, and we detect anomalies at test time.

Since the features from the standard pre-trained deep models were `rediscovered' to be effective for the task \cite{reiss2021panda}, 
recent 
studies employing them have achieved 
higher and higher performances on existing public benchmarks. Those features are proved to be representative enough for local image regions, even without any adaptation to the anomaly detection datasets at hand \cite{reiss2021panda,roth2022towards}. While visual unsupervised AD appears to be a solved problem due to these successes, attention has also been paid to more challenging problems, such as few-shot AD \cite{sheynin2021hierarchical,huang2022registration} and developing a unified model that can detect anomalies for multiple different classes of objects/textures \cite{you2022unified}. However, these studies continue to consider the one-class setting, assuming a perfect set of anomaly-free samples to be available, which usually needs manual annotation by manufacturing experts.

In this paper, we consider yet another scenario of unsupervised AD, which does not need 
any human annotations. Specifically, we consider the problem where we are not given the knowledge of normal samples for training; we want to detect anomalies in an input set of samples that might contain both normal and anomalous samples. Note that traditional machine learning often calls this setting `unsupervised AD' and the above one `semi-supervised AD' \cite{zong2018deep,akcay2019ganomaly}, unlike recent studies in computer vision. For clarity, we call the setting blind anomaly detection (BAD) in this paper. 
The recent unsupervised AD methods mentioned above are not designed for BAD and cannot directly be applied to the problem.

We then consider BAD a local patch outlier detection problem and introduce PatchCluster, which does not require human annotation and could automatically detect anomalies under BAD settings. We make the assumption that normal local features follow dense distributions and have small distances from each other in the feature space. Under this assumption, we propose to use local patch features to implicitly estimate the local feature distribution and use the average distance as the abnormality score. Unlike previous patch distribution modeling-based methods that assume features in the same spacial location follow the same distribution \cite{defard2021padim}, we cluster local patch features for the same \textit{contextual} location by distance-based nearest neighbor searching.  

We show the effectiveness of PatchCluster and prove our assumption through comprehensive experiments and analysis. Reveling the strict restriction of spacial location-based modeling, PatchCluster is robust to spacial translation and rotation. PatchCluster achieves 95.7\% and 95.9\% average image-level and pixel-level anomaly scores on MVTec AD dataset without reaching any normal training data.

\section{Related Work}
One-class anomaly detection, also known as one-class novelty detection \cite{ruff2018deep,perera2019ocgan}, is a long-standing problem in computer vision. Prior arts mainly focus on image-level outlier detection where the anomalous samples follow distributions of other semantic categories, \textit{e.g.}, detecting dog images for the cat category. Representation learning-based methods that could effectively learn the global contextual information are employed to tackle this problem, of which deep Autoencoders(AE) \cite{pidhorskyi2018generative,bergmann2018improving} and Generative Adversarial Networks(GANs) \cite{perera2019ocgan,sabokrou2018adversarially} are popular choices.

In industrial manufacturing scenarios, however, anomalies will generally occur in confined areas on a specific kind of product, making the anomalous samples very close to the normal data distribution and the task more challenging. Recent one-class anomaly detection benchmarks \cite{bergmann2019mvtec,bergmann2022beyond,bergmann2021mvtec} providing normal real-world industrial products for training draw lots of research attention and lots of attempts are paid to utilizing ImageNet \cite{krizhevsky2017imagenet} pre-trained models that could extract representative features and conduct industrial anomaly detection in a local patch feature-based manner. PaDiM \cite{defard2021padim} explicitly models the feature distribution at each spatial location. However, the assumption that patch features at the same spatial location follow the same distribution is too strict. SPADE \cite{cohen2020sub} creates a feature memory bank from the normal training data and assigns anomaly scores to test image features by kNN search. The image-level anomaly scores are still based on the global image distances. PatchCore \cite{roth2022towards} propose to use locally aware features to retain more contextual information and further utilize the greedy search method to reduce the size of the memory bank. There are also many approaches based on top of the pre-trained features that try to transfer the knowledge of normal features to a student network \cite{bergmann2020uninformed,bergmann2022anomaly,deng2022anomaly} or estimate the distribution of normal features by flow-based methods \cite{yu2021fastflow,rudolph2022fully,gudovskiy2022cflow}. 

While recent SOTA methods have shown nearly perfect anomaly detection performance on the 
public benchmark, MVTec AD dataset \cite{bergmann2019mvtec}, \textit{e.g.}, PatchCore-Ensemble achieves a 99.6\% image-level AUROC score on the MVTec AD, to the best of our knowledge, no attention has been paid to exploring achieving comparable performance without any human annotations. The one-class unsupervised anomaly detection setting requires specialists to annotate a set of normal images for training, especially for various industrial products. Under the BAD setting, most of the learning-based methods cannot be utilized directly. It should be noted that although some works have studied anomaly detection from noisy data \cite{jiang2022softpatch,chen2022deep}, \textit{i.e.}, anomaly-free training set contaminated by wrongly labeled anomalous samples, they are still under the one-class setting and need image labels from the human.

The methods most related to ours are PaDiM \cite{defard2021padim} and PatchCore \cite{roth2022towards}. We estimate the local patch feature distributions by kNN search, without explicitly modeling the distribution and revealing the overly strict assumptions of PaDiM. PatchCore only uses the nearest neighbor for anomaly scoring, which is sensitive enough under the one-class setting, however, we show that there may be several candidates that are too close to the anomalous feature in the memory bank and significantly decrease the anomaly detection sensitivity.

\begin{table}[t]
\caption{Statistical overview of the proposed blind anomaly detection settings on the MVTec AD. We report the mean value over the 15 categories.}
\centering
\resizebox{\linewidth}{!}{
\begin{tabular}{c|cccc}
\hline
 &
  Normal images &
  Anomaly images &
  \begin{tabular}[c]{@{}c@{}}Percentage of\\ normal images\end{tabular} &
  \begin{tabular}[c]{@{}c@{}}Percentage of \\ anomalous pixels\end{tabular} \\ \hline
\rotatebox[origin=c]{90}{  Mix  }  & 273 & 83 & 77\% & 1.1\% \\ \hline
\rotatebox[origin=c]{90}{  Test  } & 31  & 83 & 27\% & 3.3\% \\ \hline
\rotatebox[origin=c]{90}{  Ano  }  & 0   & 83 & 0\%  & 4.7\% \\ \hline
\end{tabular}}
\label{stat_mvtec}
\end{table}

\begin{figure*}
\centering

\includegraphics[width=0.8\linewidth]{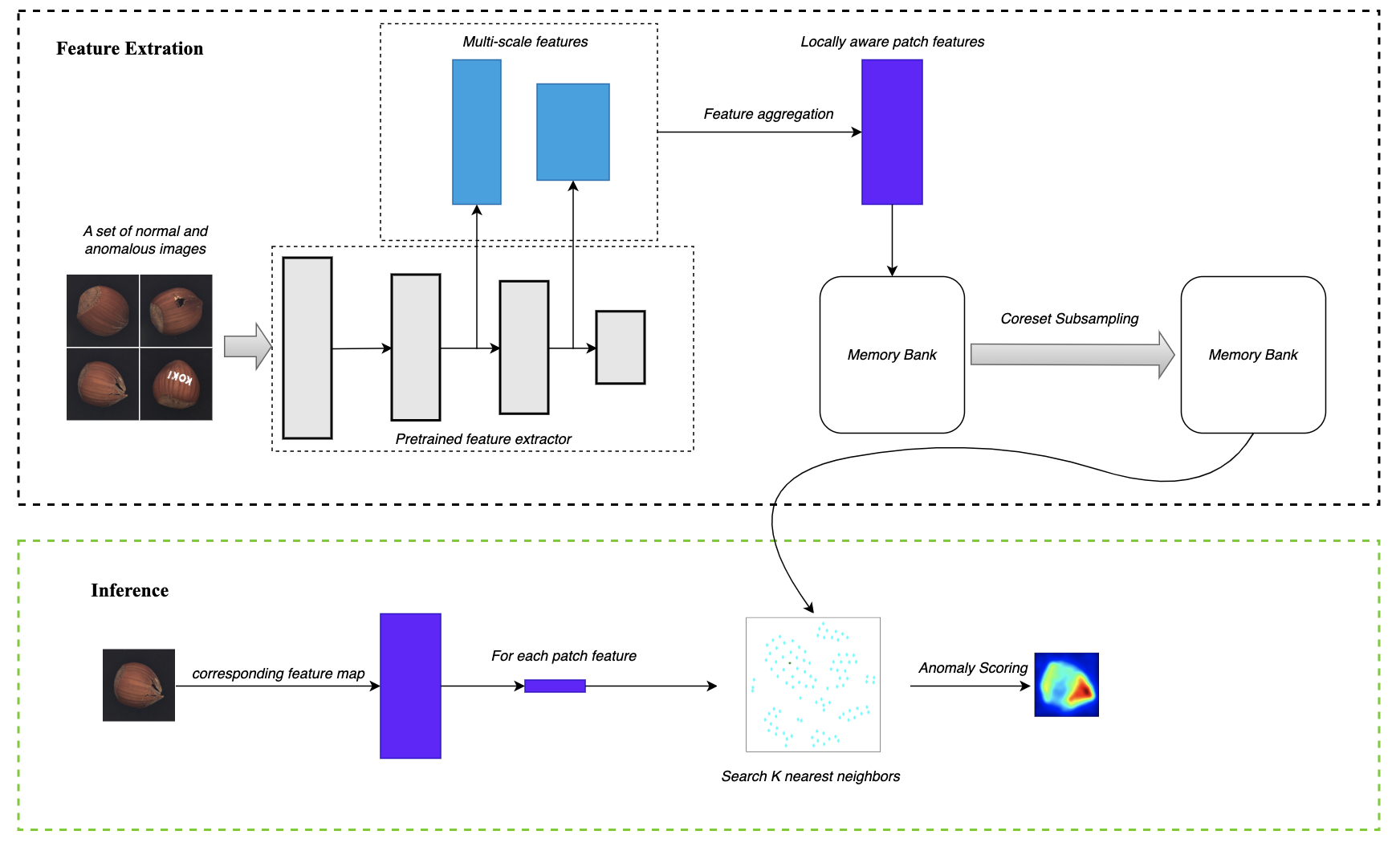}

\caption{Overview of the proposed PatchCluster. Given a mixed set of normal images and anomalous images, we use a pretrained feature extractor to extract features at multiple scales. The multi-scale features for each image are then aggregated into one locally aware peach feature map and added to the memory bank. An optional coreset subsampling process could be utilized to reduce the size of the memory bank. During inference, the $K$ nearest neighbors for one patch feature are clustered to estimate the local patch feature distribution. Without explicitly modeling the distribution, the average distance to the $K$ neighbors is used as the anomaly score.}
\label{bad_overview}
\end{figure*}

\section{Blind Anomaly Detection (BAD)}
We propose the task of blind anomaly detection (BAD), which involves identifying anomalies from a set of unknown images that may be either normal or anomalous. Given a set of images $\mathcal{S} = {I_1, I_2, ..., I_n }$, the objective is to identify anomalous images and localize the defective regions in each image. In this task, we do not use any image- and pixel-level annotations for training; we are not given any knowledge of normal or abnormal samples. Thus, recent AD methods can not be applied directly to the BAD task.

As there is currently no comprehensive real-world dataset that aligns with this task, we formulate several variations of the BAD task using the existing unsupervised anomaly detection benchmark, MVTec AD dataset. The MVTec AD dataset \cite{bergmann2019mvtec} comprises 5 texture categories and 10 object categories, and is designed for one-class anomaly detection. It includes 3629 normal images for training and 1725 normal and anomalous images for testing. We have formulated three BAD settings: MVTec-Mix, MVTec-Test, and MVTec-Ano. The MVTec-Mix setting merges the training and test images for each category, while the MVTec-Test setting is the original test dataset. The MVTec-Ano setting is the most extreme case, in which it includes only the anomalous images from the test dataset.

We present a statistical overview of the proposed BAD settings in Table \ref{stat_mvtec}. 
Under the BAD scenarios, it is not necessary to build separate training and testing splits, as opposed to supervised learning and one-class unsupervised learning. In the Ano setting, there are no normal images. Given this, the question arises whether it is possible to not only classify the two classes of images but also determine which class is normal, even though all images are anomalous. In other words, is it possible to assign low anomaly scores to normal images, even if they account for a small portion, and assign high anomaly scores to anomalous images, even though all the images contain anomalies?

Considering the specific scene, industrial manufacturing lines, we could make use of the prior knowledge that most anomalies of industrial products will occur in subtle areas, then for one anomaly image, it is more likely that the majority of the pixels are still normal. We also show the mean percentage of anomalous pixels over 15 categories in Table \ref{stat_mvtec}. The prior knowledge that the normal pixels will always account for the majority makes it possible for blind anomaly detection.

\section{Local Outlier Detection}
We then introduce the proposed PatchCluster for BAD. It is an extension of existing feature memory bank-based methods and assigns anomaly scores to each patch feature by local patch feature clustering, which makes it suitable for either one-class setting or blind anomaly detection.

\subsection{Revisiting Memory Bank-based Methods}
Under the one-class setting, building a memory bank using mid-scale features extracted by a deep pre-trained model and then applying the simple nearest neighbor search for anomaly scoring could achieve nearly perfect anomaly detection performance. We also build a feature memory bank first as the extracted features have been shown to be representative enough for anomaly detection.

Given an image $I_i$ from the dataset $\mathcal{S}$, the $l$-th layer of the pre-trained model $E$ extracts feature map $F_i^l \in \mathbb{R}^{W^l \times H^l \times C^l}$ for the image. The feature vector $f_i^l(w, h) \in \mathbb{R}^{C^l}$ at spacial location $(w, h)(w = 1, ..., W^l, h = 1, ..., H^l)$ is employed as the image patch representation.The pre-trained model could effectively extract features from low levels to high semantic levels. As the low-level features may be too generic and the high-level features are source-domain biased. The multi-scale features are concatenated along the channel dimension in PatchCore to get one single feature representation $f_i(w, h)$ at each location.  

To increase the effective receptive field of the pre-trained features while avoiding introducing ImageNet-biased knowledge, PatchCore further employs adaptive average pooling to fuse the feature vector with its neighboring features in the feature map. 

The full feature memory bank $\mathcal{M}$ is then created for all the images in the dataset $\mathcal{S}$

\begin{equation}
\begin{split}
    \mathcal{M} &= \{f_i(w, h)\}, \\
    i \in \{1, 2, ..., n\}, w &\in \{1, ..., W^l\}, h\in \{1, ..., H^l\}
\end{split}
\end{equation}

To reduce the size of the memory bank, SPADE creates a memory bank for one test image using its nearest images in the dataset, while PatchCore subsamples the memory bank into a coreset by greedy subsampling.

\subsection{Proposed PatchCluser Method}

We extend our method from the feature memory bank-based arts. We first create a memory bank $\mathcal{M}$ following PatchCore. SPADE and PatchCore score a patch feature by the distance with its nearest neighbor feature from the memory bank. It is effective enough, as there are only normal features in  $\mathcal{M}$. However, in BAD where the memory bank contains both normal and anomalous features especially $\mathcal{M}$ is built on top of the dataset $\mathcal{S}$ itself, \textit{it may also be easy to find similar patch features for the anomalous patches}. 

To address the above issue, we make the assumption that 1). features that follow the same distribution have smaller distances in the feature space and they could be used to estimate the local feature distribution in a specific \textit{contextual} location. 2). anomalies are random events and the distribution of each type of anomaly has larger variances compared to normal features. Under the above assumption, if we interpret the creating process of memory bank $\mathcal{M}$ as a sampling problem, it is possible to estimate the normal local distributions using the local features as they are similar to each other and have high probabilities to be sampled, while it is more difficult to estimate the anomalous distributions as they have more variants and have less probability to be sampled. We then build a local feature gallery $\mathcal{G}_i(h, w)$ by searching the corresponding $K$ nearest neighbors. Fig. \ref{bad_overview} gives an overview of the proposed PatchCluster.

Then the anomaly detection for each patch feature is converted to a local outlier detection problem. Under the assumption, normal local distributions are in a dense feature space while anomalous local distributions are relatively sparse. We then estimate the feature abnormality by the mean distance to its neighbors without explicitly modeling the distributions
\begin{equation}
    a_{w, h} = \frac{1}{K}\sum_{j=1}^{K}\mathrm{dist}(f_i(w, h), f_j)),\  j \in \mathcal{G}_i(h, w) 
\end{equation}

With a proper $K$, for normal patch features, it should be easy to search for enough neighbors that follow the same dense distribution. However, for an anomalous patch feature, on the one hand, the neighbors tend to have larger distances with it. On the other hand, it is also likely to fail to find enough close neighbors from $\mathcal{M}$. Consequently, after scoring all the patch features in an image, we get an anomaly score map for it. As the features are down-sampled compared to the input images, we up-sample the score map by bi-linear interpolation. Following the popular choice, to remove local noises, we apply a Gaussian filter with a kernel size of 4 to get the final anomaly score map.

We could simply choose the maximum score $a_i^*$ in a score map to account for the image-level anomaly score. We find it still robust for BAD to increase the image anomaly score if the nearest feature $f^*$ in $M$ of the corresponding patch feature $f_i^*$ has a large anomaly score

\begin{equation}
    a_i = (1 - \frac{\exp{(a_i^*)}}{\sum_{f \in \mathcal{N}(f^*)}\exp{\mathrm{dist}(f_i^*, f)}})\cdot a_i^*
\end{equation}

Where $\mathcal{N}(f^*)$ is a set of nearest neighboring patch features for $f^*$. The image anomaly scoring function is the same as PatchCore. 

\begin{figure}
  \centering
    {Input \hspace{35pt}  GT \hspace{15pt} PatchCore-25\% \hspace{10pt} Ours}
  \begin{subfigure}{1\linewidth}
    \includegraphics[width=1\linewidth]{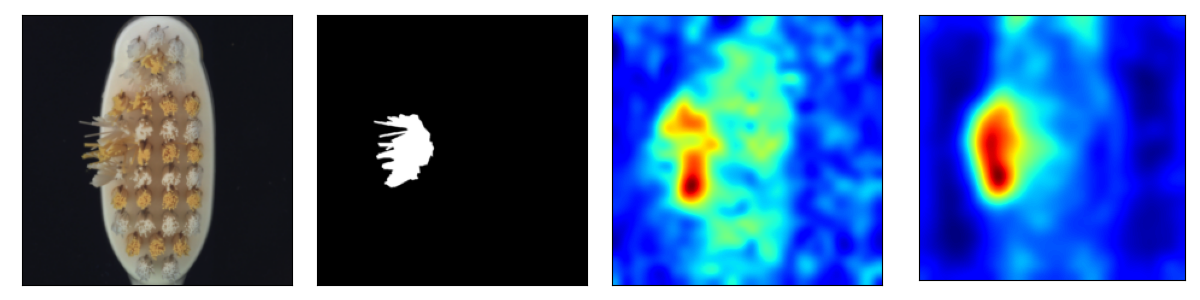}
  \end{subfigure}
    \hfill
  \begin{subfigure}{1\linewidth}
    \includegraphics[width=1\linewidth]{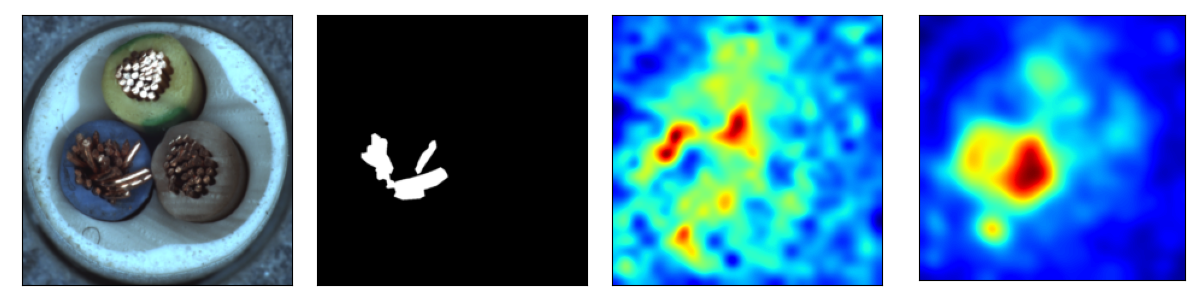}
  \begin{subfigure}{1\linewidth}
    \includegraphics[width=1\linewidth]{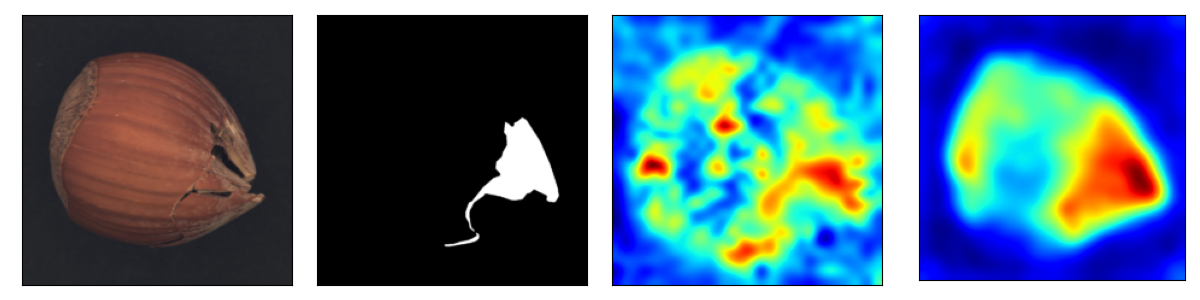}
  \end{subfigure}
    \hfill
  \end{subfigure}
\caption{Visualization examples from the toothbrush, cable, and hazelnut categories.}
\label{bad_object}
\end{figure}

\section{Experimental Results and Analysis}
\paragraph{Evaluating Metrics} We use AUROC score and PRO score to evaluate the pixel-wise anomaly detection results. The two metrics are threshold-free and the PRO metric treats each anomalous region equally.  We also use AUROC score to evaluate the image-level anomaly detection results except for the MVTec-Ano, as there is only one kind of anomalous image in this setting.

\paragraph{Implementation Details} We resize the images into a $256 \times 256$ resolution and then center crop the images into $224 \times 224$ throughout our experiments. For a fair comparison, we use the same ImageNet pre-trained WideResNet-50 \cite{zagoruyko2016wide} as the feature extractor for all evaluation methods. For PaDiM \cite{defard2021padim} and PatchCore \cite{roth2022towards}, we use the same experimental settings as in their papers. We also use the intermediate features from the output of the second and third residual stages of the feature extractor, which is the same as PatchCore. We also make SPADE follow the same choice of layers to extract features, which is different from the original SPADE that also uses the first residual stage. We find this change significantly improved the inference speed and yields better performance. As SPADE first creates a memory bank for each image by image-level nearest neighbor search, we exclude the image itself when creating the memory bank to avoid including features with high similarity for both normal and anomalous features from the image itself. For PatchCore and PatchCluster especially with memory bank subsampling, however, we aim to build a unified memory bank for all test images and cannot avoid confusing features from the image itself. We could compute the anomaly score for a given patch feature using or starting from its $k$-th nearest neighbor. However, the optimal $k$ for a set of images is highly dependent on the dataset size, the inner distribution of the images, and post-processing approaches on the memory bank such as coreset sub-sampling. To tackle this issue, we simply set $k$ to 2 to avoid searching the feature itself as the nearest neighbor. We show that using a proper $K$ value which has a clear choosing criterion makes PatchCluster robust to various BAD settings in \ref{bad_results_on_mvtec} and \ref{bad_local_feature_clustering}.

We set the number of nearest neighbors $K$ for patch feature anomaly scoring to 100 for PatchCluster-100\%, which is close to the average number of test images for each category. For the coresets with subsampling ratios of 1\%, 10\%, and 25\%, we simply reduce the $K$ to 5, 10, and 25 without careful tuning, respectively. It should be noted that as mentioned above, we do not consider the feature itself as its neighboring feature.

\begin{table*}[t]
\caption{Anomaly detection and localization performance under the blind anomaly detection settings on MVTec AD.}
\centering 
\resizebox{\linewidth}{!}{
\begin{tabular}{c|c|ccccccc}
\hline
 & Method & PaDiM & SPADE & PatchCore-LoF & PatchCore-10\% & PatchCore-25\% & PatchCore-100\% & PatchCluster-25\% \\ \hline
\multirow{3}{*}{Mix}  & Image & 95.2 & 86.8 & 87.8 & 92.7 & 86.6 & 82.9 & \textbf{97.5 }\\
                            & Pixel & 97.0 & 92.9 & 88.2 & 84.8 & 85.7 & 87.9 & \textbf{98.2} \\
                            & Pro   & \textbf{91.0} & 89.4 & 66.3 & 65.7 & 65.1 & 75.3 & \textbf{91.0} \\ \hline
\multirow{3}{*}{Test} & Image & 91.4 & 79.4 & 81.7 & 89.5 & 82.9 & 81   & \textbf{95.7} \\
                            & Pixel & 93.7 & 90.9 & 83.4 & 81   & 82.4 & 85.7 & \textbf{95.9} \\
                            & Pro   & \textbf{89.5} & 88.5 & 64.4 & 61.8 & 61.9 & 73.7 & \textbf{89.5} \\ \hline
\multirow{2}{*}{Ano} & Pixel & 92.0  & 88.6 & 80.7 & 78.1 & 81.0 & 84.5 & \textbf{94.3} \\
                            & Pro   & \textbf{88.5}  & 87.6 & 63.2 & 59.6 & 61.0 & 72.9 & 88.3 \\ \hline
\end{tabular}}
\label{bad_compare}
\end{table*}

\begin{figure}
  \centering
    {Input \hspace{35pt}  GT \hspace{15pt} PatchCore-25\% \hspace{10pt} Ours}
  \begin{subfigure}{1\linewidth}
    \includegraphics[width=1\linewidth]{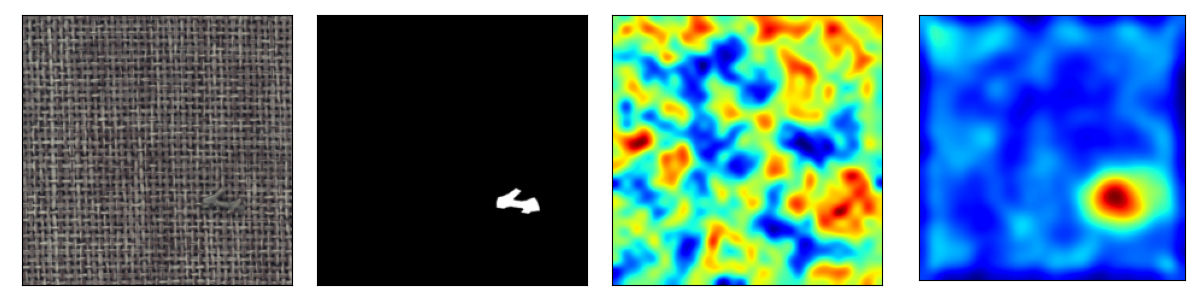}
  \end{subfigure}
    \hfill
  \begin{subfigure}{1\linewidth}
    \includegraphics[width=1\linewidth]{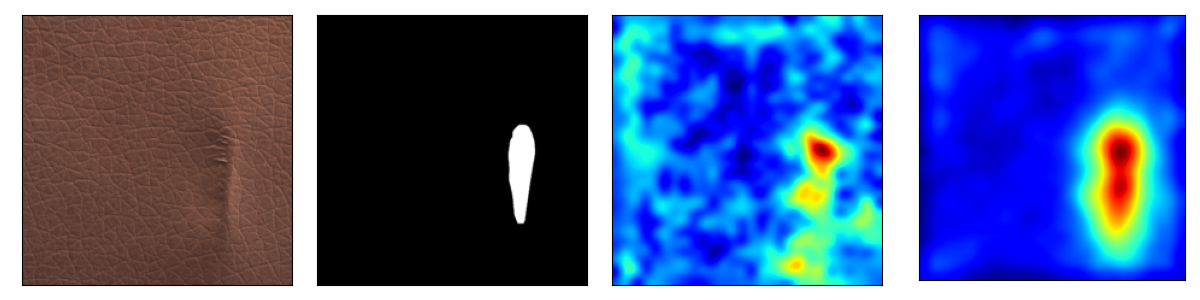}
  \begin{subfigure}{1\linewidth}
    \includegraphics[width=1\linewidth]{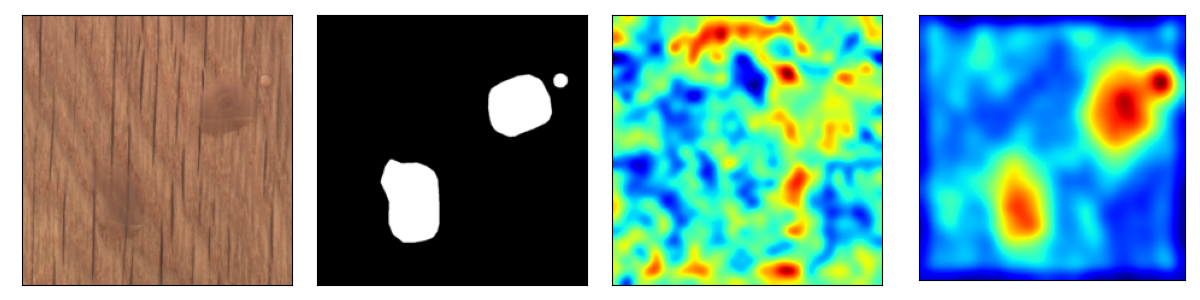}
  \end{subfigure}
    \hfill
  \end{subfigure}
\caption{Visualization examples from the carpet, leather, and wood categories.}
\label{bad_texture}
\end{figure}

\subsection{Blind Anomaly Detection on MVTec AD}
\label{bad_results_on_mvtec}
We first give an overall comparison with PaDiM, SPADE, PatchCore, and the proposed PatchCluster in Table \ref{bad_compare}. The proposed PatchCluster uses the same memory bank as PatchCore. Our proposed method outperforms all competitors by a large margin under all BAD settings regarding both image-level and pixel-level anomaly detection evaluation metrics. PatchCluster-25\% achieves 97.5\% and 98.2\% AUROC scores for anomaly detection and localization for Mix setting, which is comparable to several SOTA one-class anomaly detection approaches. Without approaching any training data under the Test setting, PatchCluster still preserves high anomaly detection performance. Under the most aggressive Ano setting, PatchCluster still achieves a 94.3\% AUROC score and a PRO score of 88.3\% for pixel-wise anomaly detection. As a comparison, 
the best image-level and pixel-level AUROC scores are only 92.7\% and 92.9\% under the Mix setting, at the influence of only 1.1\% anomalous pixels in the dataset. It should also be noted that even for the Mix setting with the smallest portion of anomalous pixels, there are still 23\% anomalous images in the dataset, which is a stringent BAD setting and practically impossible for uniform sampling from industrial product lines.

As another patch feature memory bank-based method, the modified SPADE shows better pixel-level anomaly detection than PatchCore, demonstrating the influence of confusing features from the test image itself.

PaDiM explicitly models the distributions for each fixed patch feature location. Well-modeled distributions are desired for one-class anomaly detection, however, they conflict against the performance under BAD settings as they also cover the anomalous patch features. 

We also extend the PatchCore with classical Local Outlier Factor (LoF) \cite{breunig2000lof} algorithm, a local-density-based anomaly scoring method. We use the full memory bank without coreset subsampling and calculate the relative local density for each patch feature according to the distances with its neighbors as the anomaly score. We found using 100 neighbors yield better anomaly detection results, which is consistent with our assumption.

We then report the detailed detection results of PatchCluster-25\% for each category under different BAD settings in Tab \ref{bad_detail_mvtec}. Some visualization examples under the Test setting for objects and textures are shown in Fig. \ref{bad_object} and Fig. \ref{bad_texture}, respectively. PatchCore fails to effectively assign high anomaly scores to anomalous patch features which leads to too many false-positive cases, \textit{i.e.}, normal patches could easily be detected as anomalous. PatchCluster is robust to various types of texture and object products under all BAD settings. However, it shows slightly inferior performance for categories that local spatial variations are relatively large for normal patches such as the cable, and categories with too fine-grained or large defects such as pill and transistor.

\begin{table*}[t]
\caption{Detailed anomaly detection results for each category under different settings.}
\centering
\resizebox{\linewidth}{!}{
\begin{tabular}{c|ccccccccccccccccc}
\hline
 & Metric & Bottle & Cable & Capsule & Carpet & Grid & Hazelnut & Leather & Metal Nut & Pill & Screw & Tile & Toothbrush & Transistor & Wood & Zipper & Avg. \\ \hline
\multirow{3}{*}{Mix}  & Image & 99.7 & 98.1 & 97.7 & 99.8 & 94.3 & 99.0 & 100  & 95.8 & 92.7 & 96.0 & 96.9 & 98.4 & 96.2 & 99.1 & 99.3 & 97.5 \\
                      & Pixel & 98.8 & 98.1 & 99.4 & 99.7 & 98.2 & 99.3 & 99.8 & 93.6 & 97.8 & 99.4 & 97.6 & 99.2 & 93.5 & 98.4 & 99.5 & 98.2 \\
                      & Pro   & 92.9 & 89.9 & 95.9 & 96.8 & 93.1 & 93.2 & 99.1 & 80.8 & 94.1 & 96.9 & 81.3 & 90.2 & 78.3 & 89.1 & 96.7 & 91.2 \\ \hline
\multirow{3}{*}{Test} & Image & 98.3 & 97.6 & 92.7 & 99.9 & 91.1 & 99.3 & 100  & 91.8 & 93.4 & 84.9 & 99.0 & 96.7 & 96.3 & 97.9 & 96.5 & 95.7 \\
                      & Pixel & 95.9 & 96.0 & 98.6 & 99.2 & 95.3 & 98.0 & 99.4 & 83.8 & 96.5 & 98.1 & 94.7 & 98.3 & 91.0 & 95.4 & 98.7 & 95.9 \\
                      & Pro   & 87.3 & 87.2 & 95.3 & 96.8 & 90.5 & 92.3 & 99.2 & 77.0 & 93.4 & 94.4 & 80.6 & 87.5 & 75.9 & 89.2 & 96.5 & 89.5 \\ \hline
\multirow{2}{*}{Ano}  & Pixel & 94.1 & 93.3 & 98.3 & 98.9 & 94.9 & 96.7 & 99.2 & 80.8 & 96.2 & 97.4 & 92.6 & 97.7 & 81.8 & 94.2 & 98.3 & 94.3 \\
                      & Pro   & 83.4 & 85.0 & 95.1 & 96.8 & 90.0 & 91.5 & 98.8 & 76.6 & 93.0 & 93.1 & 79.9 & 86.0 & 69.7 & 89.3 & 96.4 & 88.3 \\ \hline
\end{tabular}}
\label{bad_detail_mvtec}
\end{table*}

\begin{table*}[t]
\caption{Anomaly detection performance under the one-class setting.}

\centering
\resizebox{\linewidth}{!}{
\begin{tabular}{c|cccccccc}
\hline
Method & PaDiM & SPADE & RD & PatchCore-25\% & PatchCluster-100\% & PatchCluster-25\% & PatchCluster-10\% & PatchCluster-1\% \\ \hline
Image & 95.3 & 85.5 & 98.5          & \textbf{99.1} & 97.7 & 98.3 & 98.6          & 97.7 \\
Pixel & 97.5 & 96.0 & 97.8          & \textbf{98.1} & 97.5 & 98.0 & \textbf{98.1} & 97.5 \\
Pro   & 92.1 & 91.7 & \textbf{93.9} & 93.4          & 92.1 & 92.8 & 93.1          & 92.1 \\ \hline
\end{tabular}}
\label{bad_one_class_setting}
\end{table*}

\begin{table}[t]
\caption{Blind anomaly detection results of PatchCluster-100\% using different $K$ values.}
\centering
\resizebox{\linewidth}{!}{
\begin{tabular}{c|cccccll}
\hline
                      & K     & 1    & 10   & 50            & 100           & 150           & 200           \\ \hline
\multirow{3}{*}{Mix} & Image & 82.9 & 92.7 & \textbf{97.3} & \textbf{97.3} & 97.1          & 96.9          \\
                      & Pixel & 87.9 & 92.9 & 96.7          & 97.2          & 97.4          & \textbf{97.5} \\
                      & Pro   & 75.3 & 86.1 & 90.8          & \textbf{91.0} & \textbf{91.0} & \textbf{91.0} \\ \hline
\multirow{3}{*}{Test}  & Image & 81   & 89.8 & \textbf{95.6} & 95.3          & 94.6          & 93.9          \\
                      & Pixel & 85.7 & 90.3 & 94.3          & 94.9          & \textbf{95.1} & \textbf{95.1} \\
                      & Pro   & 73.7 & 84.1 & \textbf{89.5} & \textbf{89.5} & 89.3          & 89.0          \\ \hline
\multirow{3}{*}{Ano}  & Pixel & 84.5 & 88.6 & 92.8          & 93.3          & 93.5          & \textbf{93.6} \\
                      & Pro   & 72.9 & 83.0 & \textbf{88.5} & 88.3          & 87.9          & 87.6          \\ \hline
\end{tabular}}
\label{bad_k}
\end{table}

\begin{table}[t]
\caption{Blind anomaly detection results of SPADE-Cluster using different $K$ values.}
\centering
\resizebox{\linewidth}{!}{
\begin{tabular}{c|cccccll}
\hline
                      & K     & 1             & 10            & 20            & 50            & 100  & 150  \\ \hline
\multirow{2}{*}{Mix}  & Pixel & 92.9          & 94.3          & \textbf{94.4} & 94.2          & 93.8 & 93.5 \\
                      & Pro   & \textbf{89.4} & \textbf{89.4} & 89.0          & 87.9          & 86.5 & 85.5 \\ \hline
\multirow{2}{*}{Test} & Pixel & 90.9          & \textbf{92.4} & \textbf{92.4} & 92.1          & 91.8 & 91.4 \\
                      & Pro   & \textbf{88.5} & 88.3          & 87.7          & 86.4          & 84.9 & 83.9 \\ \hline
\multirow{2}{*}{Ano}  & Pixel & 88.6          & 89.9          & 89.9          & \textbf{90.2} & 90.0 & 89.7 \\
                      & Pro   & \textbf{87.6} & 87.2          & 86.5          & 85.1          & 83.6 & 82.6 \\ \hline
\end{tabular}}
\label{bad_spade_k}
\end{table}

\begin{table}[t]
\caption{Anomaly detection results with different coreset sub-sampling ratios.}
\centering

\begin{tabular}{c|ccccc}
\hline
                      & Ratio & 1\% & 10\% & 25\% & 100\% \\ \hline
\multirow{3}{*}{Mix}  & Image  & 90.5             & 97.2              & \textbf{97.5}     & 97.3               \\
                      & Pixel  & 95.8             & 98.0              & \textbf{98.2}     & 97.2               \\
                      & Pro    & 83.9             & 90.3              & \textbf{91.2}     & 91.0               \\ \hline
\multirow{3}{*}{Test} & Image  & 76.1             & 94.3              & \textbf{95.7}     & 95.3               \\
                      & Pixel  & 88.8             & 95.7              & \textbf{95.9}     & 94.9               \\
                      & Pro    & 72.4             & 88.6              & \textbf{89.5}     & 89.5               \\ \hline
\multirow{2}{*}{Ano}  & Pixel  & 85.8             & 93.9              & \textbf{94.3}     & 93.3               \\
                      & Pro    & 68.5             & 87.3              & \textbf{88.3}     & 88.3               \\ \hline
\end{tabular}
\label{bad_subsampling}
\end{table}

\begin{figure}
  \centering
  \begin{subfigure}{1\linewidth}
    \includegraphics[width=1\linewidth]{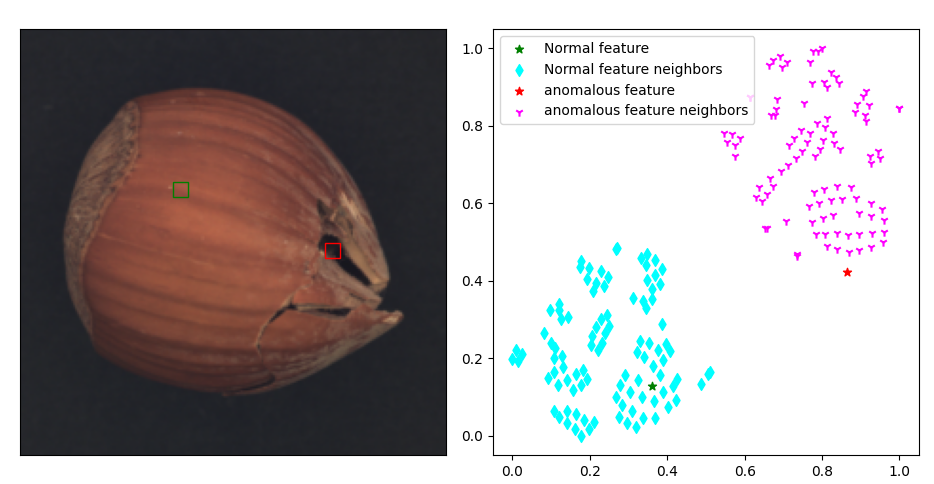}
  \end{subfigure}
    \hfill
  \begin{subfigure}{1\linewidth}
    \includegraphics[width=1\linewidth]{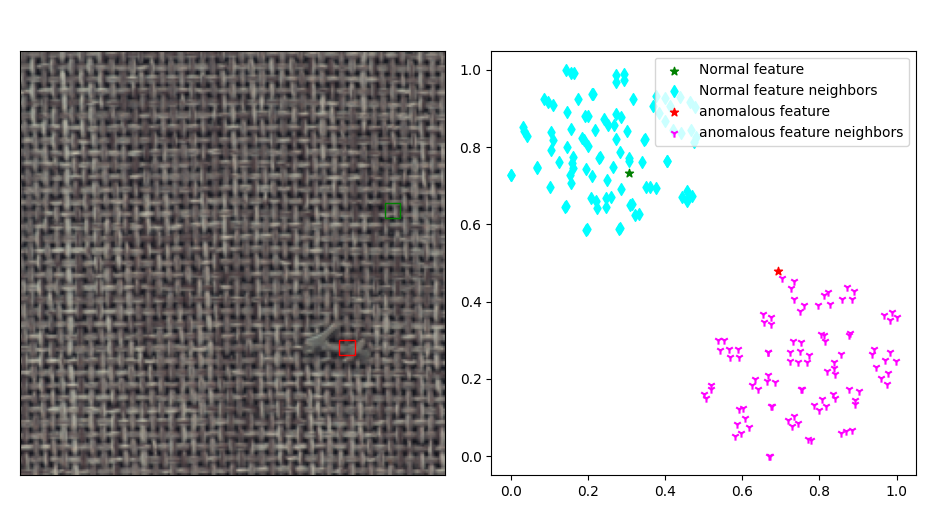}

  \end{subfigure}
\caption{Visualization of local feature clustering by nearest neighbor search.}
\label{bad_tsne}
\end{figure}

\subsection{Effectiveness of Local Feature Clustering}
\label{bad_local_feature_clustering}
We first report the performance of the proposed PatchCluster-100\% with different numbers of patches used for estimating the distance for the test patch feature and the local patch distribution in Table \ref{bad_k}. PatchCluster is stable for a wide range of $K$. It should be noted that with $K=1$ the PatchCluster is identical to PatchCore-$100\%$. The proper value of $K$ has a certain choosing criterion. From the global contextual viewpoint, each image of a certain kind of product tends to contain most kinds of local patches. Setting $K$ near the number of total images in $\mathcal{S}$ is likely to yield stable performance. We set $K$ fixed to $100$ for PatchCluster-100\% which is approximately the average number of images for each category under the Test setting. We visualize the local feature clustering under the Test setting on two examples from one object category and one texture category in Fig. \ref{bad_tsne}. The normal feature and its neighboring features tend to compromise a dense distribution and they are close in distance to each other. However, for anomalous features, there may exist several neighboring features, but most of the neighbors tend to have large distances from them.

To further analyze the effectiveness of local feature clustering, we extend SPADE into SPADE-Cluster and experimentally verify the performance. The results are shown in Table \ref{bad_spade_k}. It is also obvious that with local feature clustering, the SPADE-Cluster also shows significant improvement against SPADE. The performance of SPADE-Cluster drops with $K$ larger than$50$, which is the number of nearest images used to create the memory bank for each test image.

\subsection{Effectiveness of Memory Bank Subsampling}

Table \ref{bad_subsampling} shows the BAD performance of PatchCluster with different coreset subsampling ratios using the same greedy search method as PatchCore. We observe obvious performance drops with too small subsampling ratios, regardless of the significant memory bank size reduction and improvement of inference speed.  A similar observation for PatchCore could also be found in Table \ref{bad_compare}. We argue that the local patch-based image-level anomaly scoring methods will become unreliable if the pixel-level anomaly detection ability is worse. An underlying interpretation is the difficulty of covering not only normal patch features but also anomalous patch features that are relatively sparse in the feature space with small subsampling ratios. PatchCluster-25\% performs even better than PatchCluster-100\% as the subsampling process, to a certain degree, plays a role of a noise feature filter.

\subsection{One-Class Anomaly Detection on MVTec AD}

Table \ref{bad_one_class_setting} shows the results of PatchCluster-25\% and other methods under the one-class setting on the MVTec AD. With the patch features of the training data fully anomaly-free, the greedy-searched coreset effectively reduced the size of the memory bank while retaining strong anomaly detection ability. However, as there are no anomalous patch features in the memory bank and coreset, using a certain amount of neighboring patch features for anomaly scoring reduces the detection sensitivity, leading to slightly inferior performance compared to PatchCore.

\section{Conclusion}
We introduce the blind anomaly detection(BAD) problem for industrial inspection, a task of finding fine-grained local anomalies in a set of mixed normal and anomalous images without using any human annotations. We formulate three BAD settings, Mix, Test, and Ano based on the existing industrial anomaly detection benchmark MVTec AD dataset. 

Based on the memory bank-based method, we convert BAD as a local outlier detection problem and propose PatchCluster, a method of measuring the local patch feature's distribution using the corresponding nearest neighbors in the memory bank. The proposed approach could effectively estimate the normal feature distribution, while it fails for anomalies. PatchCluster is robust to various kinds of BAD settings and shows comparable performance with current one-class anomaly detection SOTAs.

{\small
\bibliographystyle{ieee_fullname}
\bibliography{egbib}
}

\end{document}